\documentclass{article}

\usepackage{PRIMEarxiv}

\usepackage[utf8]{inputenc} 
\usepackage[T1]{fontenc}    
\usepackage{hyperref}       
\usepackage{url}            
\usepackage{booktabs}       
\usepackage{amsfonts}       
\usepackage{nicefrac}       
\usepackage{microtype}      
\usepackage{lipsum}
\usepackage{fancyhdr}       
\usepackage{graphicx}       
\graphicspath{{media/}}     

\title{ClickGuard: Detecting and Spoiling Clickbait News with Informativeness Measures and Large Language Models
}

\author{
  Wojciech Michaluk \\
  Faculty of Mathematics and Information Science \\
  Warsaw University of Technology \\
  Warsaw, Poland \\
  \texttt{wojciech.michaluk.stud@pw.edu.pl}
  \And
  Tymoteusz Urban \\
  Faculty of Mathematics and Information Science \\
  Warsaw University of Technology \\
  Warsaw, Poland \\
  \texttt{tymoteusz.urban.stud@pw.edu.pl}
  \And
  Mateusz Kubita \\
  Faculty of Mathematics and Information Science \\
  Warsaw University of Technology \\
  Warsaw, Poland \\
  \texttt{mateusz.kubita.stud@pw.edu.pl}
  \And
  Soveatin Kuntur \\
  Faculty of Mathematics and Information Science \\
  Warsaw University of Technology \\
  Warsaw, Poland \\
  \texttt{soveatin.kuntur.dokt@pw.edu.pl}
  \And
  Anna Wróblewska \\
  Faculty of Mathematics and Information Science \\
  Warsaw University of Technology \\
  Warsaw, Poland \\
  \texttt{anna.wroblewska1@pw.edu.pl}
}

\begin{document}
\maketitle

\begin{abstract}
This paper presents an AI-driven browser extension that identifies clickbait to help users avoid misleading Internet articles. Moving beyond traditional detection, the application employs a hybrid machine learning architecture that combines transformer-based embeddings with linguistically motivated features and a custom "baitness" score. After evaluating various natural language processing techniques --- from classic vectorizers to large language model (LLM) embeddings --- an XGBoost-based model was developed that achieves an F1-score of 91\% on the open combined dataset. Most importantly, the tool can warn users before and after they access a clickbait article. After opening an article, the user receives a percentage score indicating the likelihood that it is clickbait. The prediction is explained based on the analyzed metrics, including those specifically developed within the proposed system. The browser extension also provides a clickbait spoiler --- a one- to two-sentence summary of the entire article.\\
\textbf{Demo video}: \\\href{https://www.youtube.com/watch?v=IJ1gkQV82C4}{https://www.youtube.com/watch?v=IJ1gkQV82C4}

\end{abstract}
\begin{figure}[!ht]
    \centering
    \includegraphics[width=\columnwidth]{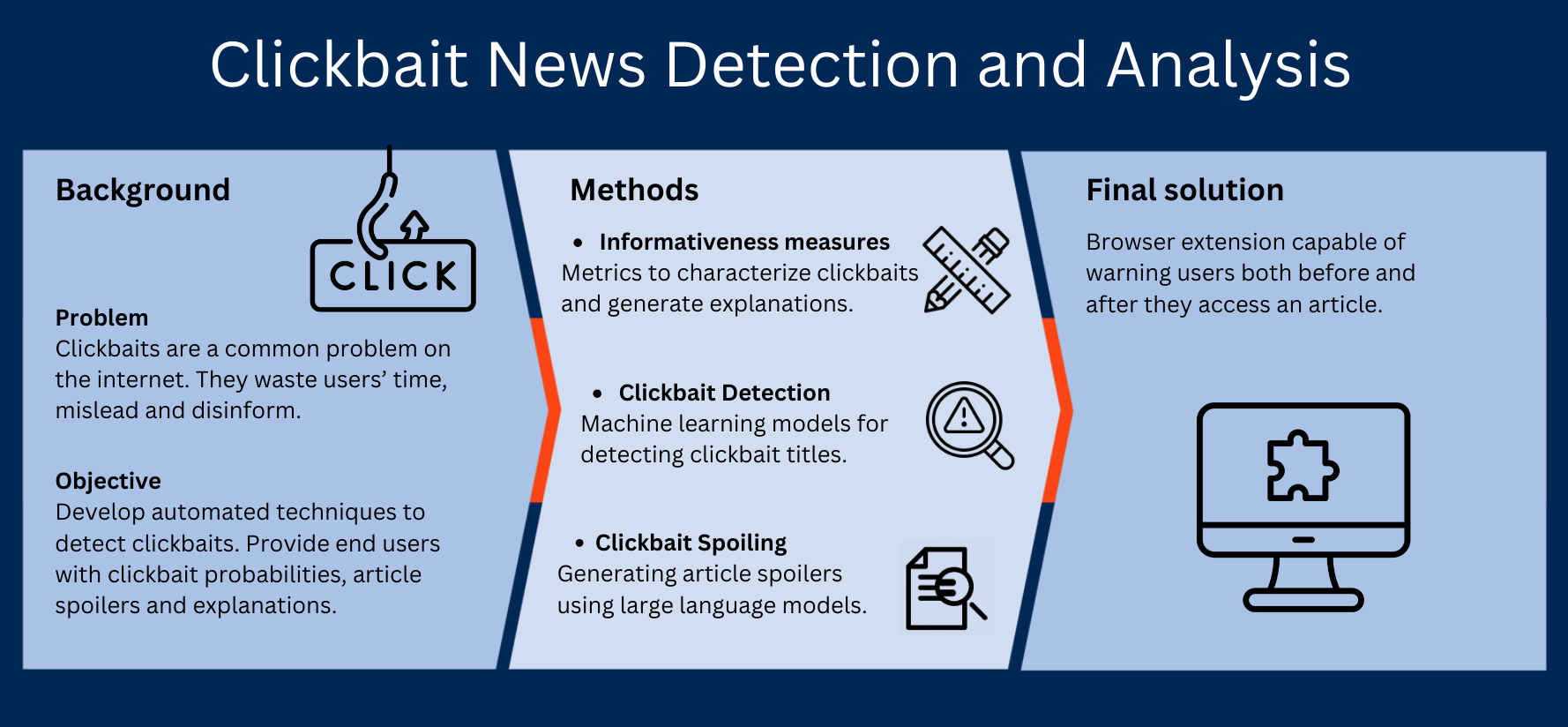}
    \caption{Visual abstract of the ClickGuard}
    \label{fig:visual_abstract}
\end{figure}


\section{Introduction}

Clickbait refers to online content specifically designed to entice readers to click a link, but that offers very little reward for doing so \cite{Scott2021Clickbait}. As a result, clickbait has become a widespread problem on the Internet, wasting users’ time and contributing to the spread of disinformation. Most existing clickbait detection approaches formulate the task as a binary classification problem and rely on full article content, multimodal inputs, or computationally expensive language models, making them largely unsuitable for real-time, real-world deployment \cite{wang2025multi} \cite{abdullah2026multimodal}. (Figure~\ref{fig:visual_abstract} presents a visual abstract summarizing the system's workflow.)

Our goal was to address this challenge by filling the gap between theoretical NLP advancements and practical, user-centric tools. We developed \textit{ClickGuard}, a browser extension that not only detects clickbait but also informs Internet users about the potential threat of clicking a clickbait article. It is worth mentioning that existing solutions often function as "black boxes". The solution presented in this article prioritizes interpretability. One implemented a hybrid machine learning framework that combines dense vector representations from Large Language Models (LLMs) with a novel set of 15 linguistically motivated "informativeness measures" --- such as the usage of superlatives, second-person pronouns, and specific punctuation patterns \cite{Yu2024MultimodalClickbait}.

A broad spectrum of architectures was evaluated, ranging from classic baselines to advanced transformer-based models such as BERT and RoBERTa \cite{roberta}. The proposed solution leverages an XGBoost classifier trained on hybrid features, achieving an F1-score of 91\% on combined open datasets for clickbait detection, and outperforming both purely embedding-based approaches and LLM-based classification. This analysis informed the selection of a model that balances high predictive accuracy with the computational efficiency required for a browser-based environment.
In addition to detection, \textit{ClickGuard} addresses the curiosity gap via automated content spoiling. Generative AI extracts and summarizes key information from clickbait articles, which allows users to satisfy their curiosity without visiting the page \cite{Hagen2022ClickbaitSpoiling} \cite{Liu_2023}.
We integrated these models into \textbf{\textit{ClickGuard}, a real-time browser extension that analyzes headlines on search engines and full articles upon navigation. Moreover, the extension provides a "baitness" score and explanations based on our informativeness metrics.} 

\begin{figure*}[t]
    \centering
    \includegraphics[width=0.8\textwidth]{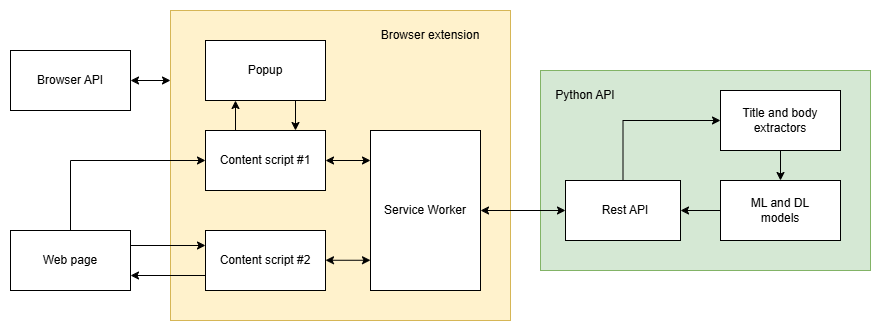}
    \caption{Client--server architecture of \textit{ClickGuard}, consisting of the browser extension and a Python REST API backend.}
    \label{fig:architecture}
\end{figure*}

\section{Related Work}
Research in clickbait detection has evolved from utilizing classic machine learning techniques~\cite{bronakowski_automatic_2023} to employing advanced deep learning architectures. Recent research shows that advanced AI models, such as Deep Recurrent Neural Networks~\cite{browser_extension_deep_rnn} and other deep learning methods~\cite {chowanda_identifying_2023}, are highly effective at detecting sensationalist headlines and clickbait. In the domain of practical applications, browser extensions such as \textit{CliNe}~\cite{clineai_chrome_extension} and \textit{ClickBaitSecurity}~\cite{browser_extension_deep_rnn} have been developed to offer real-time protection. However, most existing tools operate as binary classifiers, failing to address the user's curiosity gap. Our work bridges this limitation by incorporating generative content spoiling~\cite{hagen2022clickbaitspoilingquestionanswering} to satisfy user curiosity directly.

\section{System Architecture}
The \textit{ClickGuard} system follows a client-server architecture that offloads computationally intensive tasks from the user's browser to a scalable cloud backend --- see Figure~\ref{fig:architecture}. The solution consists of two primary components: a Chromium-based browser extension (client) and a Python microservice (server) hosted on the Google Cloud Platform.

\paragraph{Browser Extension (Client).}
Built on Chrome Manifest V3, the client consists of the following components:

\textbf{Content Script \#1 (Post-click mode):} Activated on an opened article page. It extracts the page title and body. Forwards the content to the Service Worker for backend analysis, displays the returned probability score, and spoiler in the pop-up interface.

\textbf{Content Script \#2 (Pre-click mode):} Executed on supported search engines and news pages. It scans the DOM for article links, sends URLs in batches for analysis, and injects colored risk badges next to headlines before the user clicks.

\textbf{Service Worker:} A background orchestrator responsible for internal messaging, REST API communication with the Python backend, and local caching of predictions.

\textbf{Popup Interface:} A visual front-end displaying the ''baitness'' score, interpretability explanation, and generated spoiler.
\paragraph{Backend and API.}
The backend is a containerized Python microservice (Flask) deployed on Google Cloud Run. It performs content extraction using \textit{BeautifulSoup4} and \textit{Trafilatura} and exposes a REST API with separate pre-click and post-click endpoints.

\begin{figure}[t]
    \centering
    \includegraphics[width=\columnwidth]{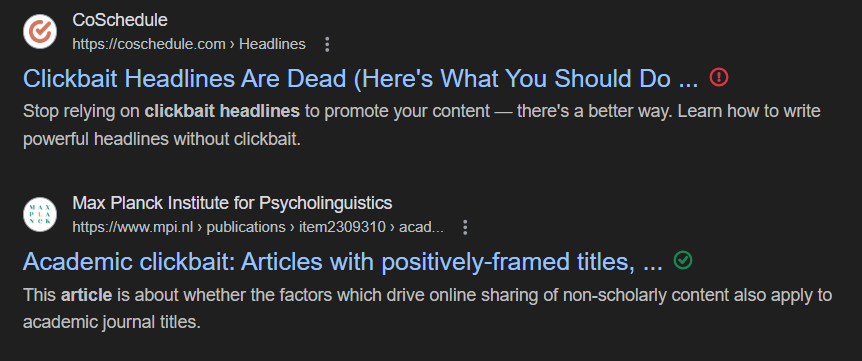}
\caption{\textit{ClickGuard} operating in pre-click mode. Icons next to headlines represent the clickbait probability calculated by a hybrid XGBoost model.}
    \label{fig:pre_warning}
\end{figure}

\textbf{Pre-click Endpoint:} Batch-analyzes URLs from search results and returns probability scores used to render colored risk badges (Figure~\ref{fig:pre_warning}).

\begin{figure}
    \centering
    \includegraphics[width=\columnwidth]{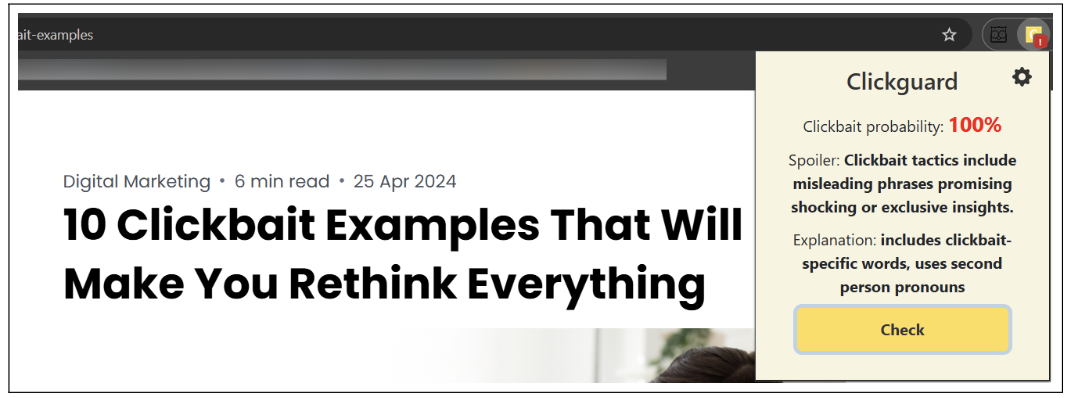}
    \caption{\textit{ClickGuard} in post-click mode with clickbait spoiling}
    \label{fig:postclick}
\end{figure}

\textbf{Post-click Endpoint:} Processes the opened article and returns a prediction, probability score, explanation, and generated spoiler (Figure~\ref{fig:postclick}).

\section{Machine Learning Models}
To enable real-time intervention against clickbait within a browser extension, one developed a two-stage pipeline: a hybrid classification model for detection and a generative model for content spoiling. One evaluated several methods ranging from traditional lexical baselines to large language models (LLMs) to balance accuracy with latency constraints.

\paragraph{Training Dataset.}
One used four English-language datasets to develop and evaluate our detection and spoiling models: \textbf{Kaggle Datasets:} Two datasets totaling approximately 53,000 news headlines with binary clickbait labels \cite{kaggle_1_source} \cite{kaggle_2_source}.
\textbf{Clickbait Challenge 2017:} 38,830 social media posts~\cite{clickbait_challenge_2017}.
\textbf{SemEval-2023 Task 5:} Approximately 4,000 entries annotated with both abstractive and extractive spoilers, used for the spoiling task~\cite{pan_webis_spoiling}.

For detection, one combined the Kaggle and Clickbait Challenge datasets and created a balanced subset of 20,000 clickbait and 20,000 non-clickbait headlines to prevent classifier bias~\cite{imbalanced_data}.

\subsection{Clickbait Detection}
For this binary classification task, one benchmarked various baselines, including TF-IDF, Word2Vec, and RoBERTa. Our top results came from a \textbf{hybrid architecture} that combines deep semantic embeddings with explicit linguistic features \cite{bronakowski_automatic_2023}.

\textbf{Deep Semantic Embeddings:}
One utilized OpenAI's \texttt{text-embedding-3-large} to generate dense vector representations of headlines. An ablation study on dimensionality indicated that projecting embeddings down to 1,000 dimensions (from 3,072) maintained predictive performance while significantly reducing computational overhead \cite{reimers_2019_sentence_bert}.\\
\textbf{Informativeness Measures:} To capture stylistic nuances not always preserved in semantic embeddings, one computed 15 handcrafted linguistically motivated features~\cite{Yu2024MultimodalClickbait}. These include readability metrics (e.g., FRES), sentiment scores (polarity, subjectivity), and structural indicators like the ratio of stop words and "bait" punctuation. Crucially, one introduced a custom \textbf{"Baitness" measure}, a composite score that aggregates eye-catchingness (e.g., capitalization, numerals) and curiosity-inducing patterns, which proved effective at separating classes~\cite{Reisenbichler2022Frontiers}.

\subsubsection{Classifier Selection}
Trained tree-based classifiers, specifically Random Forest and XGBoost, on these hybrid representations. The \textbf{XGBoost} classifier, trained on the concatenated vector of 1,000-dimensional embeddings and the 115 informativeness measures, achieved the highest performance on our dataset (test split), with an F1-score of 0.909. Table~\ref{tab:confusion_matrix} details the confusion matrix for this model, showing a balanced performance with slightly higher precision in the non-clickbait class.
\begin{table}[t] 
    \centering
    \small 
    \begin{tabular}{p{3.2cm}rr} 
        \toprule
         & \multicolumn{2}{c}{Predicted Class} \\
        \cmidrule(lr){2-3}
        Actual Class & Non-Clickbait & Clickbait \\
        \midrule
        Non-Clickbait & 1,857 & 150 \\
        Clickbait     & 208  & 1,785 \\
        \bottomrule
    \end{tabular}
    \caption{Confusion matrix for the best performing model. The layout is set to match the width of the performance comparison table.}
    \label{tab:confusion_matrix}
\end{table}
Table~\ref{tab:model_summary} summarizes the performance of our proposed model against the baselines. It should be noted that this hybrid approach outperformed models that were trained only on embeddings (without stylistic features) by approximately 4.5 percentage points.
\begin{table}[t] 
    \centering
    \small 
    \begin{tabular}{p{3.2cm}lr}
        \toprule
        Feature Representation & Classifier & F1-Score \\
        \midrule
        \textit{Baselines} & & \\
        TF-IDF & Random Forest & 0.829 \\
        Word2Vec (Google News) & XGBoost & 0.829 \\
        RoBERTa (Fine-tuned) & Transformer & 0.842 \\
        OpenAI Embeddings (3072d) & XGBoost & 0.864 \\
        \midrule
        \textit{Proposed Hybrid Model} & & \\
        \textbf{OpenAI Emb. (1000d) + 15 Features} & \textbf{XGBoost} & \textbf{0.909} \\
        \bottomrule
    \end{tabular}
    \caption{Performance comparison of clickbait detection models. Numeric values are right-aligned as per guidelines.}
    \label{tab:model_summary}
\end{table}

\subsection{Clickbait Spoiling}
Our system uses a spoiling module to generate brief summaries that satisfy the reader's curiosity. We also experimented with extractive methods using RoBERTa and fine-tuning T5-large, but encountered limitations regarding output coherence and computational resources. Finally, we decided to use GPT-4o-mini via for the browser extension. 

\section{Summary and Impact}
This paper presents \textit{ClickGuard}, a practical AI application designed to address challenges arising from the digital attention economy and online disinformation.
\textit{ClickGuard} warns users about potential manipulative content --- so-called clickbaits. 
The solution integrates AI methods: 1) a hybrid architecture combining deep semantic embeddings with 15 explicit linguistic features (achieving an F1-score of 0.909 on a big open dataset); 2) an LLM-based solution for spoiling the detected potential clickbait content.

While studying the search results page, the user is shown a pre-click warning that indicates the site's clickbait probability. After opening a new site in post-click mode, the user is provided with an additional summary of the entire page --- so-called clickbait spoiler.  
Our solution can save time and reduce user information overload when reading news articles online. 

The current version was fine-tuned on English datasets; however, adapting it to other languages or multilingual settings requires minor adjustments to the linguistic features and fine-tuning of our machine learning models. Our solution is also cost-effective because of simple language statistics as features and a small embedding model, with only additional spoiling handled by the LLM.

\section*{Acknowledgement}
This work was supported by the European Union under the Horizon Europe grant "Overcoming Multilevel Information Overload" (OMINO, \url{https://ominoproject.eu}, grant no. 101086321) and by the Polish Ministry of Education and Science within the framework of the program titled International Projects Co-Financed. However, the views and opinions expressed are those of the authors only and do not necessarily reflect those of the European Union or the European Research Executive Agency. Neither the European Union nor the European Research Executive Agency can be held responsible for them. We are thankful to PhD Daniel Dan, PhD Robert Paluch and M.Sc. Eng. Adam Majczyk for their assistance and helpful remarks. Their input was greatly appreciated.
\bibliographystyle{unsrt}  
\bibliography{references}

\end{document}